\documentclass[final]{cvpr}

\usepackage{times}
\usepackage{epsfig}
\usepackage{graphicx}
\usepackage{amsmath}
\usepackage{amssymb}

\usepackage{color}
\usepackage{bbding}
\usepackage{array,multirow,textcomp}
\usepackage{rotating}
\usepackage[skip=0.5\baselineskip]{caption}
\usepackage{makecell}
\usepackage{booktabs}
\usepackage{adjustbox}
\usepackage{array}
\usepackage{balance}
\usepackage{wrapfig}

\usepackage{tabularx}

\newcolumntype{Y}{>{\Centering\arraybackslash}X}

\makeatletter
\@namedef{ver@everyshi.sty}{}
\makeatother
\usepackage{tikz}
\usetikzlibrary{shapes.geometric}
\tikzstyle{connector} = [->, thick, rounded corners]
\definecolor{myblue}{HTML}{2cb1e1}
\definecolor{mypred}{rgb}{0.82,0.2,0.2}
\definecolor{myloss}{rgb}{0.6,0.8,0.1}
\definecolor{myyellow}{rgb}{1.0,0.8509803921568627,0.1843137254901961}

\usepackage{algpseudocode}
\usepackage{algorithm}

\newcolumntype{C}[1]{>{\centering\let\newline\\\arraybackslash\hspace{0pt}}m{#1}}

\newcommand*\rot{\rotatebox{90}}
\def\quads{\hskip0.5em\relax}
\newcommand{\SB}[1]{\textbf{#1}}
\newcommand{\IT}[1]{\textit{#1}}

\newcommand{\MC}[1]{\mathcal{#1}}
\newcommand{\rpm}{\raisebox{.3ex}{$\scriptstyle\pm$}}
\newcolumntype{R}[2]{%
    >{\adjustbox{angle=#1,lap=\width-(#2)}\bgroup}%
    l%
    <{\egroup}%
}

\newcommand\blfootnote[1]{%
  \begingroup
  \renewcommand\thefootnote{}\footnote{#1}%
  \addtocounter{footnote}{-1}%
  \endgroup
}

\newcommand*\rots{\multicolumn{1}{R{60}{1em}}}

\usepackage[pagebackref=true,breaklinks=true,colorlinks,bookmarks=false]{hyperref}

\def\cvprPaperID{6105}
\def\confYear{CVPR 2021}

\def\toptitlebar{
  \hrule height4pt
  \vskip .25in
}

\def\bottomtitlebar{
  \vskip .25in
  \hrule height1pt
  \vskip .25in
}

\begin{document}

\title{Learning to Relate Depth and Semantics for Unsupervised Domain Adaptation}
\date{}
\author{
    Suman Saha* \\
    ETH Zurich
    \and
    Anton Obukhov* \\
    ETH Zurich
    \and
    Danda Pani Paudel \\
    ETH Zurich
    \and
    Menelaos Kanakis \\
    ETH Zurich
    \and
    Yuhua Chen \\
    ETH Zurich
    \and
    Stamatios Georgoulis \\
    ETH Zurich
    \and
    Luc Van Gool \\
    ETH Zurich, KU Leuven
}

\maketitle

\begin{abstract}
We present an approach for encoding visual task relationships to improve model performance in an Unsupervised Domain Adaptation (UDA) setting. 
Semantic segmentation and monocular depth estimation are shown to be complementary tasks; in a multi-task learning setting, a proper encoding of their relationships can further improve performance on both tasks.
Motivated by this observation, we propose a novel Cross-Task Relation Layer (CTRL), which encodes task dependencies between the semantic and depth predictions.
To capture the cross-task relationships, we propose a neural network architecture that contains task-specific and cross-task refinement heads.
Furthermore, we propose an Iterative Self-Learning (ISL) training scheme, which exploits semantic pseudo-labels to provide extra supervision on the target domain.
We experimentally observe improvements in both tasks' performance because the complementary information present in these tasks is better captured.
Specifically, we show that:
(1) our approach improves performance on all tasks when they are complementary and mutually dependent;
(2) the CTRL helps to improve both semantic segmentation and depth estimation tasks performance in the challenging UDA setting;
(3) the proposed ISL training scheme further improves the semantic segmentation performance.
The implementation is available at 
\href{https://github.com/susaha/ctrl-uda}{https://github.com/susaha/ctrl-uda}.
\end{abstract}

\begin{figure}[t!]
\begin{center}
   \includegraphics[width=\linewidth]{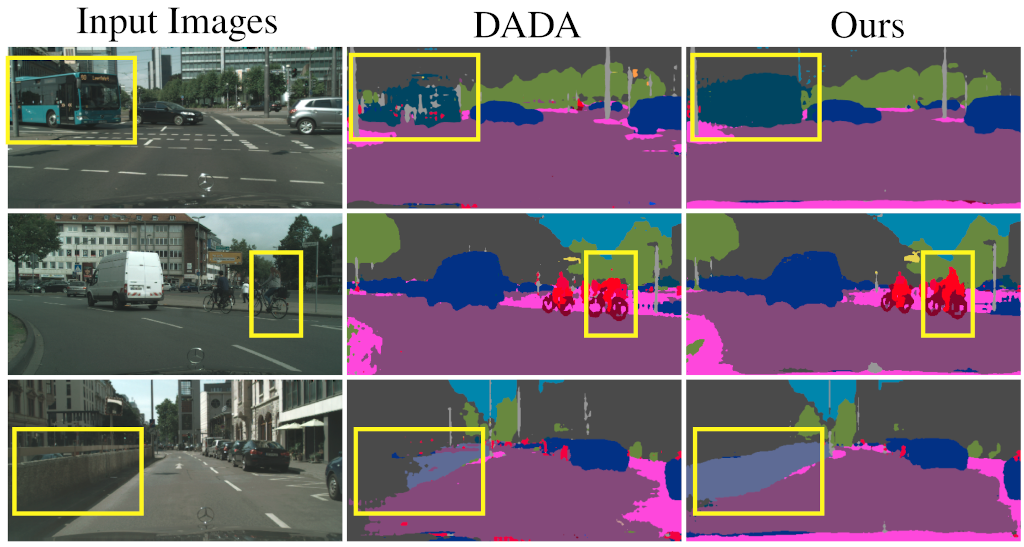}
\end{center}
\caption{
Semantic segmentation improvement with our approach to unsupervised domain adaptation over the state-of-the-art DADA~\cite{vu2019dada} method. Left to right: Cityscapes test images, DADA, and the proposed method (CTRL). Our model correctly segments the ``bus'', ``rider'', and ``wall'' classes underrepresented in the target domain (highlighted).
}
\label{fig:introteaser}
\end{figure}

\section{Introduction}
\label{sec:intro}
\blfootnote{Corresponding author: Suman Saha (\href{mailto://suman.saha@vision.ee.ethz.ch}{suman.saha@vision.ee.ethz.ch})}%
\blfootnote{* Equal contribution.}%
Semantic segmentation and monocular depth estimation are two important computer vision tasks that allow us to perceive the world around us and enable agents' reasoning, e.g., in an autonomous driving scenario. Moreover, these tasks have been shown to be complementary to each other, i.e., information from one task can improve the other task's performance~\cite{kendall2018multi,maninis2019attentive,vandenhende2020mti}.
Domain Adaptation (DA)~\cite{csurka2017comprehensive} refers to maximizing model performance in an environment with a smaller degree of supervision (the \textit{target domain}) relative to what the model was trained on (the \textit{source domain}). Unsupervised Domain Adaptation (UDA) assumes only access to the unannotated samples from the target domain at train time -- the setting of interest in this paper, explained in greated detail in Sec.~\ref{sec:udarecap}.

Recent domain adaptation techniques~\cite{lee2018spigan,vu2019dada} proposed to leverage depth information available in the source domain to improve semantic segmentation on the target domain. 
However, they lack an explicit multi-task formulation to relate depth and semantics, that is to say, \IT{how each semantic category relates to different depth levels}. The term \IT{depth levels} refers to different discrete ranges of depth values, i.e., ``near'' (1-5m); ``medium-range'' (5-20m), or ``far'' ({$>$}20m).
This paper aims to design a model that learns explicit relationships between different visual semantic classes and depth levels within the UDA context.

To this end, we design a network architecture and a new multitask-aware feature space alignment mechanism for UDA. 
First, we propose a Cross-Task Relation Layer (CTRL) -- a novel parameter-free differentiable module tailored to capture the task relationships given the network's semantic and depth predictions. 
Second, we utilize a Semantics Refinement Head (SRH) that explicitly captures cross-task relationships by learning to predict semantic segmentation given predicted depth features.
Both CTRL and SRH boost the model's ability to effectively encode correlations between semantics and depth, thus improving predictions on the target domain. 
Third, we employ an Iterative Self Learning (ISL) scheme. Coupled with the model design, it further pushes the performance of semantic segmentation. 
As a result, our method achieves state-of-the-art semantic segmentation performance on three challenging UDA benchmarks (Sec.~\ref{sec:exp}).
Fig.~\ref{fig:introteaser} demonstrates our method's effectiveness by comparing semantic predictions of classes underrepresented in the target domain to predictions made by the previous state-of-the-art method.

The paper is organized as follows: 
Sec.~\ref{sec:relwork} discusses the related work; 
Sec.~\ref{sec:method} describes the proposed approach to UDA, the network architecture, and the learning scheme;
Sec.~\ref{sec:exp} presents the experimental analysis with ablation studies;
Sec.~\ref{sec:conclusion} concludes the paper.

\section{Related Work}
\label{sec:relwork}

\textbf{Semantic Segmentation.} refers to the task of assigning a semantic label to each pixel of an image. Conventionally, the task has been addressed using hand-crafted features combined with classifiers, such as Random Forests~\cite{shotton2008semantic}, SVMs~\cite{fulkerson2009class}, or Conditional Random Fields~\cite{ladicky2010and}.
Powered by the effectiveness of Convolutional Neural Networks (CNNs)~\cite{lecun1998gradient}, we have seen an increasing number of deep learning-based models. Long~et~al.~\cite{long2015fully} were among the first to use fully convolutional networks (FCNs) for semantic segmentation. Since then, this design has quickly become a state-of-the-art method for the task. The encoder-decoder design is still widely used~\cite{yu2015multi,chen2018encoder,badrinarayanan2017segnet,zhao2017pyramid, chen2017deeplab}.

\textbf{Cross-domain Semantic Segmentation.} Training deep networks for semantic segmentation requires large amounts of labeled data, which presents a significant bottleneck in practice, as acquiring pixel-wise labels is a labor-intensive process. A common approach to address the issue is to train the model on a source domain and apply it to a target domain in a UDA context. However, this often causes a performance drop due to the domain shift. 
Domain Adaptation aims to solve the issue by aligning the features from different domains. DA is a highly active research field, and techniques have been developed for various applications, including image classification~\cite{ganin2015unsupervised,li2017deeper,long2015learning,lu2017unsupervised}, object detection~\cite{chen2018domain}, fine-grained recognition~\cite{gebru2017fine}, etc.

More related to our method are several works on unsupervised domain adaptation for semantic segmentation~\cite{zhang2017curriculum,sankaranarayanan2017unsupervised,zou2018unsupervised,chen2018road,vu2019advent, iqbal2020mlsl, yang2020label, zhou2020uncertainty, paul2020domain, yang2020context}. This problem has been tackled with curriculum learning~\cite{zhang2017curriculum}, GANs~\cite{sankaranarayanan2017unsupervised}, adversarial training on the feature space~\cite{chen2018road}, output space~\cite{tsai2018learning}, or entropy maps~\cite{vu2019advent}, self-learning using pseudo- or weak labels \cite{zou2018unsupervised, paul2020domain, iqbal2020mlsl}. 
However, prior works typically only consider adapting semantic segmentation while neglecting any multi-task correlations. 
A few methods~\cite{chen2019learning,vu2019dada} model correlations between semantic segmentation and depth estimation, similarly to our work, yet -- as explained in Sec.~\ref{sec:intro} -- these works come with crucial limitations.

\textbf{Monocular Depth Estimation.} 
Similar to semantic segmentation, monocular depth estimation is dominated by CNN-based methods~\cite{Eigen2014,fu2018deep,laina2016deeper,li2015depth}. \cite{Eigen2014}~introduced a CNN-based architecture for depth estimation, which regresses a dense depth map. Their approach was then improved by incorporating techniques such as a CRF~\cite{Liu:2016,li2015depth} and multi-scale CRF techniques~\cite{xu2017multi}. Besides, improvements in the loss design itself also lead to better depth estimation. Examples include the reverse Huber (berHu) loss~\cite{owen2007robust,zwald2012berhu}, and the ordinal regression loss~\cite{fu2018deep}.

\textbf{Multi-task Learning for Semantic Segmentation and Depth Estimation.} Within the context of multi-task learning, semantic segmentation is shown to be highly correlated with depth estimation, and vice versa~\cite{zamir2018taskonomy,xu2018pad,kendall2018multi,zhang2018joint,zhang2019pattern,maninis2019attentive,standley2019tasks,vandenhende2020mti,vandenhende2020revisiting,kanakis2020reparameterizing}. To leverage this correlation, some authors have proposed to learn them jointly~\cite{ramirez2018geometry,jiao2018look,chen2019towards}. In particular,~\cite{neven2017fast,jiao2018look,vandenhende2019branched,bruggemann2020automated} proposed to share the encoder and use multiple decoders, whereas a shared conditional decoder is used in~\cite{chen2019towards}. Semantic segmentation was also demonstrated to help guide the depth training process~\cite{guizilini2020semantically,jiang2019sense}. 

In this paper, we build upon these observations. We argue that task relationships, like the ones between depth and semantics, are not entirely domain-specific. As a result, if we correctly model these relationships in one domain, they can be transferred to another domain to help guide the DA process. The proposed method and its components are explicitly designed around this hypothesis.

\begin{figure*}[t!]
\begin{center}
\resizebox{\linewidth}{!}{
  \input{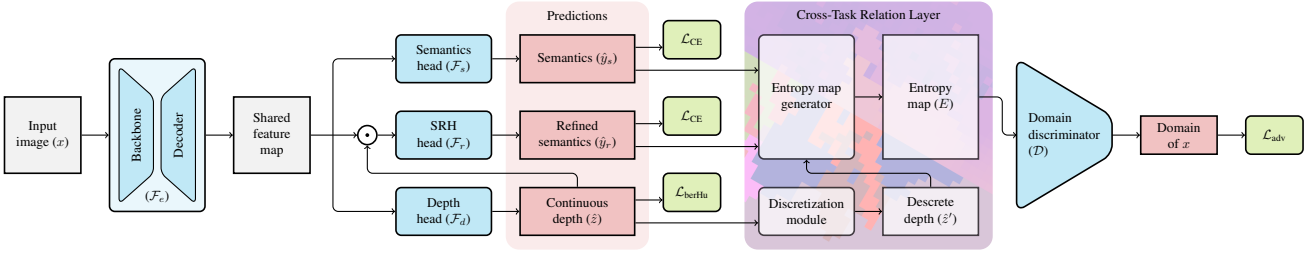}
}
\end{center}
\caption{
Overview of the proposed neural architecture (Sec.~\ref{sec:problemformulation}) and the CTRL module (Sec.~\ref{sec:ctrl}). Supervised losses (in the middle) are applied only on the source domain; the rest of the data flow is domain-agnostic.
Legend: \textcolor{myblue}{learned modules}, \textcolor{mypred}{predictions}, \textcolor{myloss}{loss functions}; rounded corners denote operators, rectangles denote activations.
}
\label{fig:overview}
\end{figure*}

\section{Method}
\label{sec:method}

In this section, we describe our approach to UDA in the autonomous driving setting.
Sec.~\ref{sec:overview} presents an overview of the proposed approach;
Sec.~\ref{sec:problemformulation} explains the notation and problem formulation;
Sec.~\ref{sec:supervisedlearning} describes supervision on the source domain;
Sec.~\ref{sec:ctrl} presents the CTRL module design; 
Sec.~\ref{sec:isl} describes the ISL technique;
Sec.~\ref{sec:nnarch} prescribes the rest of the network architecture details.

\subsection{Overview} 
\label{sec:overview}

The primary hypothesis behind our approach is that task dependencies persist across domains, i.e., most semantic classes fall under a finite depth range. We can exploit this information from source samples and transfer it to target using adversarial training.
As our goal is to train the network in a UDA setting, we follow an adversarial training scheme~\cite{hoffman2016fcns,tsai2018learning} to learn domain invariant representations. 

Unlike~\cite{vu2019dada} that directly aligns a combination of semantics and depth features, we wish to design a joint feature space for domain alignment by fusing the task-specific and the cross-task features and then learn to minimize the domain gap through adversarial training. To this end, we propose CTRL -- a novel module that constructs the joint feature space by computing entropy maps of both the semantic label and discretized depth distributions (Fig.~\ref{fig:overview}). Thus, CTRL entropy maps, generated on the source and target domains, are expected to carry similar information. 

Further enhancement of semantic segmentation performance appears possible by utilizing the Iterative Self-Learning (ISL) training scheme, which does not require expensive patch-based pseudo-label generation like~\cite{iqbal2020mlsl}.
As our CTRL helps the network to predict high-quality predictions (Fig.~\ref{fig:introteaser}), ISL training exploits high-confidence predictions as supervision (pseudo-labels) on the target domain.

\subsection{Problem Formulation}
\label{sec:problemformulation}

Let $\mathfrak{D}^{(s)}$ and $\mathfrak{D}^{(t)}$ denote the source and target domains, with samples from them represented by tuples $(x^{(s)}, y^{(s)}, z^{(s)})$ and $(x^{(t)})$ respectively, where $x \in \mathbb{R}^{H \times W \times 3}$ are color images, $y \in \{1, . . ., C\}^{H \times W}$ are semantic annotations with $C$ classes, and $z \in [Z_\text{min}, Z_\text{max}]^{H \times W}$ are depth maps from a finite frustum. Furthermore, $\mathcal{F}_e$ is the shared feature extractor, which includes a pretrained backbone, and a decoder; $\mathcal{F}_s$ and $\mathcal{F}_d$ are the task-specific semantics and depth heads, respectively; $\mathcal{F}_r$ is the SRH (Fig.~\ref{fig:overview}). 

First, $\mathcal{F}_e$ extracts a shared feature map to be used by SRH and task-specific semantics and depth heads.
The semantics head $\mathcal{F}_s$ predicts a semantic segmentation map $\hat{y}_s = \mathcal{F}_s(\mathcal{F}_e(x))$ with $C$ channels per pixel, denoting predicted class probabilities.
The depth head $\mathcal{F}_d$ predicts a real-valued depth map $\hat{z} = \mathcal{F}_d(\mathcal{F}_e(x))$, where each pixel is mapped into the finite frustum specified in the source domain.
We further employ SRH to learn the cross-task relationship between semantics and depth by making it predict semantics from the shared feature map, attenuated by the predicted depth map.
Formally, the shared feature map is point-wise multiplied by the predicted depth map, and then SRH predicts a second (auxiliary) semantic segmentation map: $\hat{y}_r = \mathcal{F}_r(\hat{z} \odot \mathcal{F}_e(x))$.

We refer to the part of the model enclosing the $\mathcal{F}_e, \mathcal{F}_s, \mathcal{F}_r, \mathcal{F}_d$ modules as a prediction network. 
The predictions made by the network on the source and target domains are denoted as $(\hat{y}_s^{(s)}, \hat{y}_r^{(s)}, \hat{z}^{(s)})$ and $(\hat{y}_s^{(t)}, \hat{y}_r^{(t)}, \hat{z}^{(t)})$, respectively.
We upscale these predictions along the spatial dimension to match the original input image dimension $H \times W$ before any further processing. 
Given these semantics and depth predictions on the source and target domains,
we optimize the network cost using supervised loss on the source domain, and unsupervised domain alignment loss on the target domain within the same training process.

\subsection{Supervised Learning}
\label{sec:supervisedlearning}
Since the semantic segmentation predictions $\hat{y}_s^{(s)}$, $\hat{y}_r^{(s)}$ and ground truth $y^{(s)}$ are represented as pixel-wise class probabilities over $C$ classes, we employ the standard cross-entropy loss with the semantic heads:
\begin{equation}
\label{eq:ce}
    \MC{L}_\text{CE}(\hat{y}, y) = -\sum_{i=1}^C y_i \log \hat{y}_i.
\end{equation}
We use the berHu loss (the reversed Huber criterion~\cite{laina2016deeper}) for penalizing depth predictions:
\begin{equation}
\begin{split}
\MC{L}_\text{berHu}(\hat{z}, z)&=
\left\{
\begin{array}{ll}
|z-\hat{z}|                     & |z-\hat{z}| \leq L,\\
\frac{(z-\hat{z})^2 + L^2}{2L}  & |z-\hat{z}| > L,
\end{array}
\right. \\
L&=0.2 \max(|z-\hat{z}|).
\end{split}
\end{equation}
Following~\cite{kendall2018multi}, we regress inverse depth values (normalized disparity), which is shown to improve the precision of predictions on the full range of the view frustum. 
The parameters of the network $\theta_e$, $\theta_s$, $\theta_r$, $\theta_d$ (parameterizing $\mathcal{F}_e$, $\mathcal{F}_s$, $\mathcal{F}_r$, $\mathcal{F}_d$ modules), collectively denoted as $\theta_{\text{net}}$, are learned to minimize the following supervised objective on the source domain:
\begin{equation}
\begin{split}
\label{eq:sup_optimization_cost}
    \min_{\theta_{\text{net}}} \mathop{\mathbb{E}}_{\mathfrak{D}^{(s)}}  
    \big[ & \MC{L}_\text{CE}(\hat{y}_s^{(s)}, y^{(s)}) + \lambda_r \MC{L}_\text{CE}(\hat{y}_r^{(s)}, y^{(s)}) \; + \\
        & \lambda_d \MC{L}_\text{berHu}(\hat{z}^{(s)}, z^{(s)}) \big]
\end{split}
\end{equation}
where $\lambda_{r}$ and $\lambda_{d}$ are the hyperparameters weighting relative importance of the SRH and depth supervision.

\subsection{Cross-Task Relation Layer}  
\label{sec:ctrl}
In the absence of ground truth annotations for the target samples, we train the network on the target images using unsupervised domain alignment loss.
Existing works either align source and target domain in a semantic space~\cite{vu2019advent} or a depth-aware semantic space~\cite{vu2019dada} by fusing the continuous depth predictions with predicted semantic maps.
Here, we argue that simple fusion of the continuous depth prediction into the semantics does not enable the network to learn useful semantic features at different depth levels. Instead, explicit modeling is required to achieve this goal.

Humans learn to relate semantic categories at each discrete depth level differently. For example, ``sky'' is ``far away'' (large depth), ``vehicles'' are ``nearby'', ``road'' appears to be both ``far'' and ``nearby''.
Taking inspiration from the way humans relate semantic and depth,
we design a CTRL (Fig. \ref{fig:overview}) that captures the semantic class-specific 
dependencies at different discrete depth levels.
Moreover, CTRL also preserves task-specific information by fusing task-specific and task-dependent features learned by the semantics, depth, and refinement (SRH) heads. 
CTRL consists of a depth discretization, an entropy map generation, and a fusion layer described in the following subsections.

\subsubsection{Depth Discretization Module}
The prediction made by the depth head $\hat{z}$ contains continuous depth values. We want to map it to a discrete probability space to learn visual semantic features at different depth levels. 
We quantize the view frustum depth range into a set of representative discrete values following the spacing-increasing discretization (SID)~\cite{fu2018deep}.
Such discretization assigns progressively large depth sub-ranges further away from the point of view into separate bins, which allows us to simulate the human perception of depth relations in the scene, with a finite number of categories.

Given the depth range $[Z_\text{min}, Z_\text{max}]$ and the number of depth bins $K$, SID outputs a $K$-dimensional vector of discretization bin centers $b$ as follows:
\begin{equation}
    b_i = Z_\text{min}^{1 - (2i + 1)/2K} \cdot Z_\text{max}^{(2i + 1)/2K},\quad i=0,\dots,K{-}1
\end{equation}
We can now assign probabilities of the predicted depth values falling into the defined bins:
\begin{equation} 
\label{eq:dep_disc}
    \hat{z}' = \text{softmax}(-(\hat{z} - b)^2).
\end{equation}

\subsubsection{Joint Space for Domain Alignment}
The task-dependency $\hat{y}_{r}$ (output by SRH), alongside the task-specific semantics $\hat{y}_{s}$ and depth $\hat{z}'$ probability maps, can be considered as discrete distributions over semantic classes and depth levels. As we do not have access to the ground truth labels for the target domain, one way to train the network to predict high-confidence predictions is by minimizing the uncertainty (or entropy) in the predicted distributions over the target domain~\cite{vu2019advent}.
The source and target domains share similar spatial features, and it is recommended to align them in the structured output space~\cite{hoffman2018cycada}. 

To this end, we propose a novel UDA training scheme, where task-specific and task-dependent knowledge is transferred from the source to the target domain by constraining the target distributions to be similar to the source by aligning the entropy maps of $\hat{y}_{r}$, $\hat{y}_{s}$, and $\hat{z}'$.
Note that unlike \cite{vu2019dada, vu2019advent}, which constrain only on the task-specific space ($\hat{y}_{s}$ in our case) for domain alignment, we train the network to output highly certain predictions by aligning features in the task-specific and task-dependent spaces.

We argue that aligning source and target distributions jointly in task-specific and task-dependent spaces helps to bridge the domain gap for underrepresented classes, which are learned poorly without the presence of a joint representation.
To encode such a joint representation, we generate entropy maps as follows:
\begin{equation} \label{eq:entropy_maps}
\begin{split}
\begin{gathered}
    \mathcal{E}(p) = -p \odot \log p \\
    E_r = \mathcal{E}(\hat{y}_r), \; E_s = \mathcal{E}(\hat{y}_s), \; E_d = \mathcal{E}(\hat{z}').
\end{gathered}
\end{split}
\end{equation}
We then concatenate these maps along the channel dimension to get the fused entropy map $E = \text{concat}(E_r, E_s, E_d)$ and employ adversarial training on it.

For aligning the source and target domain distributions, we train the proposed segmentation and depth prediction network (parameterized by $\theta_\text{net}$) and the discriminator network $\mathcal{D}$ (parameterized by $\theta_{\mathcal{D}}$) following an adversarial learning scheme.
More specifically, the discriminator is trained to correctly classify the sample domain being either source or target given only the fused entropy map: 
\begin{equation} \label{eq:disc_loss}
\begin{split}
    \min_{\theta_{\mathcal{D}}} \big\{
    &\mathop{\mathbb{E}}_{\mathfrak{D}^{(s)}} 
    \big[ \log \mathcal{D}(E^{(s)}) \big] + \\
    &\mathop{\mathbb{E}}_{\mathfrak{D}^{(t)}} 
    \big[ \log \big( 1 - \mathcal{D}(E^{(t)}) \big) \big] \big\}
\end{split}
\end{equation}
At the same time, the prediction network parameters are learned to maximize the domain classification loss (i.e., fooling the discriminator) on the target samples using the following optimization objective:
\begin{equation} \label{eq:disc_loss_2}
    \min_{\theta_{\text{net}}} 
    \mathop{\mathbb{E}}_{\mathfrak{D}^{(t)}}
    \left[ \log \mathcal{D}(E^{(t)}) \right]
\end{equation}

We use the hyperparameter $\lambda_\text{adv}$ weighing the relative importance of the adversarial loss (\ref{eq:disc_loss_2}).
Our training scheme jointly optimizes the model parameters of the prediction network ($\theta_{\text{net}}$) and the discriminator ($\theta_{\mathcal{D}}$).
Updates to the prediction network and the discriminator happen upon every training iteration; however, when updating the prediction network, the discriminator parameters are kept fixed. Parameters of the discriminator are updated separately using the domain classification objective (Eq.~\ref{eq:disc_loss}).

\subsection{Iterative Self Learning}
\label{sec:isl}
Following prior work \cite{zou2018unsupervised}, we train our network end-to-end using an ISL scheme using Algorithm 1.
We first train the prediction ($\theta_\text{net}$) and discriminator ($\theta_{\mathcal{D}}$) networks for $Q_{1}$ iterations.
We then generate semantic pseudo-labels ($\widetilde{y}^{(t)}$) on the target training samples $x^{(t)}$ using the trained prediction network.

We further train the prediction network on the target training samples using pseudo-labels supervision and a masked cross-entropy loss (Eq.~\ref{eq:ce}), masking target prediction pixels with confidence less than $0.9$, for $Q_3$ iterations.
Instead of training the prediction network using SL only once, we iterate over generating high-confidence pseudo-labels
and self-training $Q_2$ times to refine the pseudo-labels, further resulting in better quality semantics output on the target domain.

We show in the ablation studies (Sec.~\ref{sub:exp_ablation}) that our ISL scheme outperforms the simple SL.
The discriminator network parameters ($\theta_{\mathcal{D}}$) are kept fixed during self-training.

\begin{algorithm} 
 \caption{ISL$(\mathfrak{D}^{(s)},\mathfrak{D}^{(t)}, \theta_\text{net}, \theta_{\mathcal{D}})$}
 \begin{algorithmic}[1]
 \State Train prediction ($\theta_\text{net}$) and discriminator ($\theta_{\mathcal{D}}$) networks on source and target domains for $Q_1$ iterations;
 \For{$Q_2$ times}
 \State Generate $\widetilde{y}_s^{(t)} = \mathcal{F}_s(\mathcal{F}_e(x^{(t)}))$ using trained $\theta_\text{net}$;
 \State Train $\theta_\text{net}$ on $(x^{(t)}, \widetilde{y}_s^{(t)})$ for $Q_3$ iterations;
 \EndFor
 \end{algorithmic}
\end{algorithm}

\subsection{Network Architecture}
\label{sec:nnarch}
The shared part of the prediction network $\mathcal{F}_e$ consists of a ResNet-101 backbone and a decoder (Fig. \ref{fig:overview}).
The decoder consists of four convolutional layers; 
its outputs are fused with the backbone output features, which
are denoted as the ``shared feature map''.
This shared feature map is then fed forward to the respective semantics and semantics refinement heads.
Following the residual auxiliary block \cite{mordan2018revisiting} (as in \cite{vu2019dada}), we place the depth prediction head between the last two convolutional layers of the decoder.
In the supplementary materials, we show that our proposed approach is not sensitive to the residual auxiliary block and performs equally well with a standard multi-task learning network architecture (i.e., a shared encoder followed by multiple task-specific decoders).
We adopt the Deeplab-V2~\cite{chen2017deeplab} architectural design with Atrous Spatial Pyramid Pooling (ASPP) for the prediction heads.
We use DC-GAN~\cite{radford2015unsupervised} as our domain discriminator for adversarial learning. 

\section{Experiments}
\label{sec:exp}

\begingroup
\setlength{\tabcolsep}{2pt}
\renewcommand{\arraystretch}{1.3}
\begin{table*}[ht]
\normalsize
\centering
\caption{
Semantic segmentation performance (IoU and mIoU, \%) comparison to the prior art.
All models are trained and evaluated using the EP1 protocol. mIoU* is computed on a subset of $13$ classes, excluding those marked with *. For our method, we report the results of the run giving the best mIoU, as well as 68\% confidence interval over five runs as \textit{mean{\rpm}std}.
}
\resizebox{\linewidth}{!}{
\begin{tabular}{l @{\quad} c @{\quad} cccccccccccccccc @{\quad} c @{\quad} c}
\toprule 
&&\multicolumn{18}{c}{SYNTHIA $\rightarrow$ Cityscapes (16 classes)} \\
\cmidrule{3-20}
Models & Depth & \rots{road} & \rots{sidewalk\quads\quads} & \rots{building} & \rots{wall*} & \rots{fence*} & \rots{pole*} & \rots{light} & \rots{sign} & \rots{veg} & \rots{sky} & \rots{person} & \rots{rider} & \rots{car} & \rots{bus} & \rots{mbike} & \rots{bike} & mIoU $\uparrow$ & mIoU* $\uparrow$\\
\midrule
SPIGAN-no-PI \cite{lee2018spigan} &            & 69.5 & 29.4 & 68.7 & 4.4  & 0.3 & 32.4 & 5.8  & 15.0 & 81.0 & 78.7 & 52.2 & 13.1 & 72.8 & 23.6 & 7.9  & 18.7 & 35.8 & 41.2 \\
SPIGAN       \cite{lee2018spigan} & \checkmark & 71.1 & 29.8 & 71.4 & 3.7  & 0.3 & \SB{33.2} & 6.4  & \SB{15.6} & 81.2 & 78.9 & 52.7 & 13.1 & 75.9 & 25.5 & 10.0 & 20.5 & 36.8 & 42.4 \\
AdaptSegnet  \cite{tsai2018learning} &            & 79.2 & 37.2 & 78.8 & -    & -   & -    & 9.9  & 10.5 & 78.2 & 80.5 & 53.5 & 19.6 & 67.0 & 29.5 & 21.6 & 31.3 & -    & 45.9 \\
AdaptPatch   \cite{tsai2019domain} &            & 82.2 & 39.4 & 79.4 & -    & -   & -    & 6.5  & 10.8 & 77.8 & 82.0 & 54.9 & 21.1 & 67.7 & 30.7 & 17.8 & 32.2 & -    & 46.3 \\
CLAN         \cite{luo2019taking} &            & 81.3 & 37.0 & 80.1 & -    & -   & -    & \SB{16.1} & 13.7 & 78.2 & 81.5 & 53.4 & 21.2 & 73.0 & 32.9 & 22.6 & 30.7 & -    & 47.8 \\
Advent       \cite{vu2019advent} &            & 87.0 & 44.1 & 79.7 & 9.6  & 0.6 & 24.3 & 4.8  & 7.2  & 80.1 & 83.6 & 56.4 & \SB{23.7} & 72.7 & 32.6 & 12.8 & 33.7 & 40.8 & 47.6 \\
DADA       \cite{vu2019dada}   & \checkmark & \SB{89.2} & \SB{44.8} & 81.4 & 6.8  & 0.3 & 26.2 & 8.6  & 11.1 & 81.8 & \SB{84.0} & 54.7 & 19.3 & 79.7 & 40.7 & 14.0 & 38.8 & 42.6 & 49.8 \\
\midrule
%
\SB{Ours (best mIoU)} & \checkmark & 88.1 & 39.1 & \SB{81.8} & \SB{19.6} & \SB{0.7} & 30.2 & 7.7 & 13.8 & \SB{82.4} & 83.0 & \SB{60.2} & 22.0 & \SB{81.7} & \SB{47.6} & \SB{24.8} & 35.0 & \SB{44.9} & \SB{51.3} \\
\SB{Ours (confidence)}  
& \checkmark
& \begin{tabular}{@{}r@{}} 85.5 \\[-6pt] {\rpm}3.3 \end{tabular}  
& \begin{tabular}{@{}r@{}} 40.6 \\[-6pt] {\rpm}4.4 \end{tabular}  
& \begin{tabular}{@{}r@{}} 80.0 \\[-6pt] {\rpm}1.1 \end{tabular}  
& \begin{tabular}{@{}r@{}} \SB{16.5} \\[-6pt] \SB{{\rpm}}\SB{3.3} \end{tabular}  
& \begin{tabular}{@{}r@{}} \SB{0.73} \\[-6pt] \SB{{\rpm}}\SB{0.2} \end{tabular}  
& \begin{tabular}{@{}r@{}} 28.5\\[-6pt] {\rpm}1.6 \end{tabular}   
& \begin{tabular}{@{}r@{}} 7.2 \\[-6pt] {\rpm}3.5 \end{tabular}   
& \begin{tabular}{@{}r@{}} 11.7 \\[-6pt] {\rpm}3.4 \end{tabular}  
& \begin{tabular}{@{}r@{}} 80.7 \\[-6pt] {\rpm}1.6 \end{tabular}  
& \begin{tabular}{@{}r@{}} 82.4 \\[-6pt] {\rpm}0.6 \end{tabular}  
& \begin{tabular}{@{}r@{}} \SB{60.2} \\[-6pt] \SB{{\rpm}}\SB{1.1} \end{tabular}  
& \begin{tabular}{@{}r@{}} 22.2 \\[-6pt] {\rpm}4.6 \end{tabular}  
& \begin{tabular}{@{}r@{}} 79.1 \\[-6pt] {\rpm}5.1 \end{tabular}  
& \begin{tabular}{@{}r@{}} \SB{47.2} \\[-6pt] \SB{{\rpm}}\SB{5.7} \end{tabular}  
& \begin{tabular}{@{}r@{}} \SB{26.0} \\[-6pt] \SB{{\rpm}}\SB{4.8} \end{tabular}  
& \begin{tabular}{@{}r@{}} \SB{40.6} \\[-6pt] \SB{{\rpm}}\SB{5.6} \end{tabular}  
& \begin{tabular}{@{}r@{}} \SB{44.3} \\[-6pt] \SB{{\rpm}}\SB{0.6} \end{tabular}  
& \begin{tabular}{@{}r@{}} \SB{51.0} \\[-6pt] \SB{{\rpm}}\SB{0.7} \end{tabular} \\    
\bottomrule
\end{tabular}
}
\label{table:sc16class}
\end{table*}
\endgroup

\begingroup
\setlength{\tabcolsep}{2pt}
\renewcommand{\arraystretch}{1.3}
\begin{table*}[ht]
\normalsize
\centering
\caption{
Semantic segmentation performance (IoU and mIoU, \%) comparison to the prior art. All models are trained and evaluated using the EP2 and EP3 protocols at different resolutions, as indicated in the resolution (``Res.'') column. For our method, we report the results of the run giving the best mIoU, as well as 68\% confidence interval over five runs as \textit{mean{\rpm}std}.
}
\resizebox{\linewidth}{!}{
\begin{tabular}{c @{\quad} l  c @{\quad} ccccccc @{\quad} c c @{\quad} ccccccc @{\quad} c}
\toprule
&  &  & \multicolumn{8}{c}{(a) SYNTHIA $\rightarrow$ Cityscapes (7 classes)} && \multicolumn{8}{c}{(b) SYNTHIA $\rightarrow$ Mapillary (7 classes)}  \\
\cmidrule{4-11}\cmidrule{13-20}
Res. & Model & Depth & \rots{flat} & \rots{const} & \rots{object} & \rots{nature} & \rots{sky} & \rots{human} & \rots{vehicle\quads\quads} & mIoU $\uparrow$ && \rots{flat} & \rots{const} & \rots{object} & \rots{nature} & \rots{sky} & \rots{human} & \rots{vehicle} & mIoU $\uparrow$ \\
\midrule
\multirow{7}{*}{\rot{$320 \times 640$}} 
& SPIGAN-no-PI \cite{lee2018spigan} &            & 90.3 & 58.2 & 6.8  & 35.8 & 69.0 & 9.5  & 52.1 & 46.0 && 53.0 & 30.8 & 3.6  & 14.6 & 53.0 & 5.8  & 26.9 & 26.8 \\
& SPIGAN       \cite{lee2018spigan} & \checkmark & \SB{91.2} & 66.4 & 9.6  & 56.8 & 71.5 & 17.7 & 60.3 & 53.4 && 74.1 & 47.1 & 6.8  & 43.3 & 83.7 & 11.2 & 42.2 & 44.1 \\
& Advent      \cite{vu2019advent} &  & 86.3 & 72.7 & 12.0 & 70.4 & 81.2 & 29.8 & 62.9 & 59.4 && 82.7 & 51.8 & 18.4 & 67.8 & 79.5 & 22.7 & 54.9 & 54.0 \\
& DADA        \cite{vu2019dada} & \checkmark  & 89.6 & 76.0 & 16.3 & 74.4 & 78.3 & \SB{43.8} & 65.7 & 63.4 && 83.8 & 53.7 & \SB{20.5} & 62.1 & 84.5 & 26.6 & 59.2 & 55.8 \\
\cmidrule{2-20}
& \SB{Ours (best mIoU)} & \checkmark 
& 90.9  
& \SB{78.3}  
& \SB{16.6}  
& \SB{77.4}  
& \SB{83.6}  
& 43.5  
& \SB{67.6}  
& \SB{65.4}  
&& \SB{85.8}  
& \SB{59.8}  
& 18.4  
& \SB{71.0}  
& \SB{89.7}  
& \SB{44.7}  
& \SB{64.1}  
& \SB{61.9} \\  
& \SB{Ours (confidence)} 
& \checkmark 
& \begin{tabular}{@{}r@{}}  90.2 \\[-6pt] {\rpm}0.6 \end{tabular} 
& \begin{tabular}{@{}r@{}}  \SB{77.9} \\[-6pt] \SB{{\rpm}}\SB{0.7} \end{tabular} 
& \begin{tabular}{@{}r@{}}  16.1 \\[-6pt] {\rpm}0.5 \end{tabular} 
& \begin{tabular}{@{}r@{}}  \SB{76.5} \\[-6pt] \SB{{\rpm}}\SB{1.0} \end{tabular} 
& \begin{tabular}{@{}r@{}}  \SB{82.9} \\[-6pt] \SB{{\rpm}}\SB{0.9} \end{tabular} 
& \begin{tabular}{@{}r@{}}  42.0 \\[-6pt] {\rpm}5.0 \end{tabular} 
& \begin{tabular}{@{}r@{}}  \SB{68.7} \\[-6pt] \SB{{\rpm}}\SB{1.4} \end{tabular} 
& \begin{tabular}{@{}r@{}}  \SB{64.9} \\[-6pt] \SB{{\rpm}}\SB{0.3} \end{tabular} 
&& \begin{tabular}{@{}r@{}}  \SB{85.5} \\[-6pt] \SB{{\rpm}}\SB{0.2} \end{tabular} 
& \begin{tabular}{@{}r@{}}  \SB{58.6} \\[-6pt] \SB{{\rpm}}\SB{1.5} \end{tabular} 
& \begin{tabular}{@{}r@{}}  19.1 \\[-6pt] {\rpm}1.3 \end{tabular} 
& \begin{tabular}{@{}r@{}}  \SB{69.2} \\[-6pt] \SB{{\rpm}}\SB{2.5} \end{tabular} 
& \begin{tabular}{@{}r@{}}  \SB{88.7} \\[-6pt] \SB{{\rpm}}\SB{1.1} \end{tabular} 
& \begin{tabular}{@{}r@{}}  \SB{44.0} \\[-6pt] \SB{{\rpm}}\SB{0.9} \end{tabular} 
& \begin{tabular}{@{}r@{}}  \SB{63.6} \\[-6pt] \SB{{\rpm}}\SB{1.8} \end{tabular} 
& \begin{tabular}{@{}r@{}}  \SB{61.3} \\[-6pt] \SB{{\rpm}}\SB{0.5} \end{tabular} \\ 
\midrule
\multirow{6}{*}{\rot{Full}} 
& Advent       \cite{vu2019advent} &            & 89.6 & 77.8 & 22.1 & 76.3 & 81.4 & 54.7 & 68.7 & 67.2 && 86.9 & 58.8 & 30.5 & 74.1 & 85.1 & 48.3 & 72.5 & 65.2 \\

& DADA \cite{vu2019dada} & \checkmark & 92.3 & 78.3 & 25.0 & 75.5 & 82.2 & 58.7 & 72.4 & 69.2$^{*}$ && 86.7 & 62.1 & 34.9 & 75.9 & 88.6 & 51.1 & 73.8 & 67.6 \\

& Oracle (only-target) & & 97.6 & 87.9 & 46.0 & 87.9 & 88.8 & 69.1 & 88.6 & 80.8 && 95.0 & 84.2 & 54.8 & 87.7 & 97.2 & 70.2 & 87.5 & 82.4 \\
\cmidrule{2-20}
& \SB{Ours (best mIoU)} & \checkmark 
& \SB{93.0} 
& \SB{81.3} 
& \SB{29.0} 
& \SB{78.8} 
& \SB{83.2} 
& \SB{63.2} 
& \SB{77.5} 
& \SB{72.3} 
&& \SB{88.9} 
& \SB{66.5} 
& \SB{35.7} 
& \SB{78.9} 
& \SB{92.0} 
& \SB{61.3} 
& \SB{79.2} 
& \SB{71.8}  \\ 
& \SB{Ours (confidence)} 
& \checkmark 
& \begin{tabular}{@{}r@{}}  92.1 \\[-6pt] {\rpm}0.7 \end{tabular} 
& \begin{tabular}{@{}r@{}}  \SB{80.7} \\[-6pt] \SB{{\rpm}}\SB{0.7} \end{tabular} 
& \begin{tabular}{@{}r@{}}  \SB{28.3} \\[-6pt] \SB{{\rpm}}\SB{0.5} \end{tabular} 
& \begin{tabular}{@{}r@{}}  \SB{77.8} \\[-6pt] \SB{{\rpm}}\SB{1.7} \end{tabular} 
& \begin{tabular}{@{}r@{}}  \SB{83.1} \\[-6pt] \SB{{\rpm}}\SB{1.1} \end{tabular} 
& \begin{tabular}{@{}r@{}}  \SB{60.9} \\[-6pt] \SB{{\rpm}}\SB{3.8} \end{tabular} 
& \begin{tabular}{@{}r@{}}  \SB{76.4} \\[-6pt] \SB{{\rpm}}\SB{1.0} \end{tabular} 
& \begin{tabular}{@{}r@{}}  \SB{71.3} \\[-6pt] \SB{{\rpm}}\SB{0.9} \end{tabular} 
&&\begin{tabular}{@{}r@{}}  \SB{88.5} \\[-6pt] \SB{{\rpm}}\SB{0.3} \end{tabular} 
& \begin{tabular}{@{}r@{}}  \SB{64.8} \\[-6pt] \SB{{\rpm}}\SB{2.8} \end{tabular} 
& \begin{tabular}{@{}r@{}}  34.4 \\[-6pt] {\rpm}1.2 \end{tabular} 
& \begin{tabular}{@{}r@{}}  \SB{78.9} \\[-6pt] \SB{{\rpm}}\SB{1.0} \end{tabular} 
& \begin{tabular}{@{}r@{}}  \SB{90.5} \\[-6pt] \SB{{\rpm}}\SB{2.0} \end{tabular} 
& \begin{tabular}{@{}r@{}}  \SB{61.3} \\[-6pt] \SB{{\rpm}}\SB{1.4} \end{tabular} 
& \begin{tabular}{@{}r@{}}  \SB{78.1} \\[-6pt] \SB{{\rpm}}\SB{1.1} \end{tabular} 
& \begin{tabular}{@{}r@{}}  \SB{70.9} \\[-6pt] \SB{{\rpm}}\SB{0.7} \end{tabular} \\ 
\bottomrule
\multicolumn{19}{l}{\footnotesize{$^{*}$ The correct mean of class IoU values in Table 2 of \cite{vu2019dada}.}}
\end{tabular}
}
\label{table:scm7class}
\end{table*}
\endgroup

\subsection{UDA Benchmarks}
\label{subsec:uda_benchmarks}
We use three standard UDA evaluation protocols (EPs) to validate our model:
\SB{EP1:} SYNTHIA $\rightarrow$ Cityscapes (16 classes),
\SB{EP2:} SYNTHIA $\rightarrow$ Cityscapes (7 classes), and
\SB{EP3:} SYNTHIA $\rightarrow$ Mapillary (7 classes).
A detailed explanation of these settings can be found in~\cite{vu2019dada}. 
In all settings, the SYNTHIA dataset \cite{ros2016synthia} is used as the synthetic source domain.
In particular, we use the \texttt{SYNTHIA-RAND-CITYSCAPES} split consisting of 9,400 synthetic images and their corresponding pixel-wise semantic and depth annotations.
For target domains, we use Cityscapes \cite{cordts2016cityscapes} and Mapillary Vistas \cite{neuhold2017mapillary} datasets.
Following EP1, we train models on 16 classes common to SYNTHIA and Cityscapes;
in EP2 and EP3, models are trained on 7 classes common to SYNTHIA, Cityscapes, and Mapillary.
We use intersection-over-union to evaluate segmentation: IoU (class-IoU) and mIoU (mean-IoU).
To promote reproducibility and emphasize significance of our results, we report two outcomes: the best mIoU, and the confidence interval.
The latter is denoted as \textit{mean{\rpm}std} collected over five runs, thus describing a 68\%~confidence interval centered at \textit{mean}%
\footnote{
  Class-IoU values of the "best mIoU" setting can be less than the mean of the class confidence interval at the expense of other classes performance.
}.
For depth, we use Absolute Relative Difference ($|\text{Rel}|$), Squared Relative Difference (Rel$^2$), Root Mean Squared Error (RMS), its log-variant LRMS;
and the accuracy metrics~\cite{eigen2014depth} as denoted by $\delta_1$, $\delta_2$, and $\delta_3$. 
For each metric, we use $\uparrow$ and $\downarrow$ to denote the improvement direction.

\subsection{Experimental Setup}
\label{sec:experimentalsetup}
All our experiments are implemented in PyTorch~\cite{paszke2017automatic}.
Backbone network is a ResNet-101~\cite{he2016deep} initialized with ImageNet~\cite{deng2009imagenet} weights.
The prediction and discriminator networks are optimized with SGD~\cite{bottou2010large} and Adam~\cite{kingma2014adam} with learning rates $2.5 \times 10^{-4}$ and $10^{-4}$ respectively.
Throughout our experiments, we use $\lambda_{r} = 1.0$, $\lambda_{d} = \lambda_{adv} = 10^{-3}$.
For generating depth bins, we use $Z_\text{min}=1$m, $Z_\text{max}=655.36$m, and $K=15$.
In all ISL experiments, parameters of the algorithm are: $Q_1=65K$, $Q_2=5$, $Q_3=5K$.
Link to the project page with source code is in the Abstract.

\begin{figure}[ht!]
\begin{center}
   \includegraphics[width=1.0\linewidth]{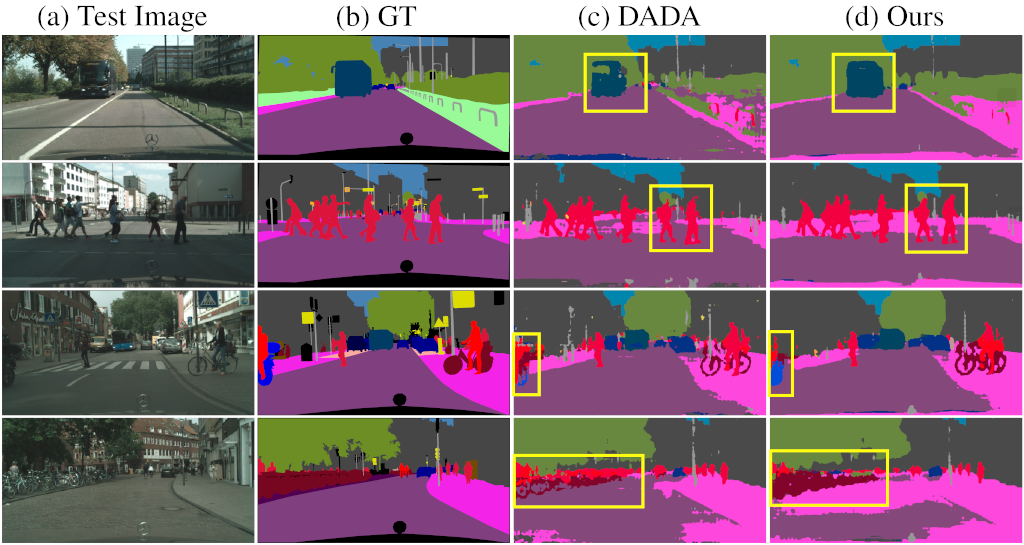}
\end{center}
\caption{
Qualitative semantic segmentation results with EP1: SYNTHIA $\rightarrow$ Cityscapes (16 classes).
(a) Images from Cityscapes validation set;
(b) ground truth annotations;
(c) DADA~\cite{vu2019dada} predictions;
(d) our model predictions.
Our method demonstrates notable improvements over \cite{vu2019dada} on ``bus'', ``person'', ``motorbike'', and ``bicycle'' classes as highlighted using the yellow boxes.
   }
\label{fig:exp_qr_1}
\end{figure}

\begin{figure}[ht!]
\begin{center}
   \includegraphics[width=1.0\linewidth]{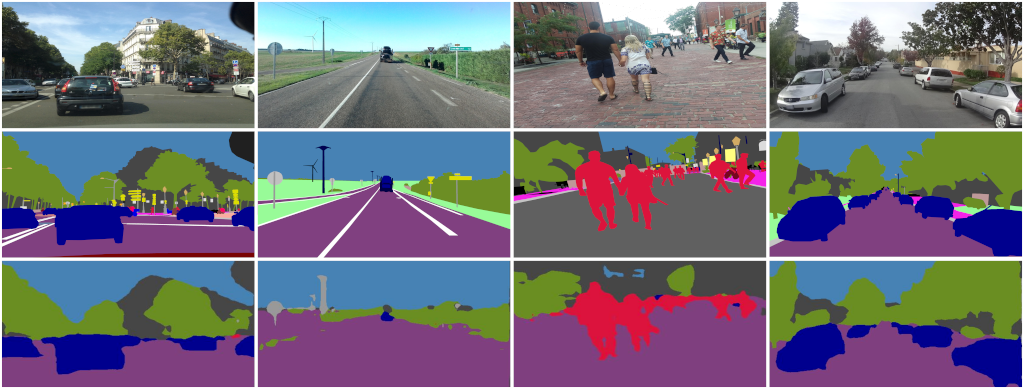}
\end{center}
   \caption{
   Qualitative semantic segmentation results with EP3: SYNTHIA $\rightarrow$ Mapillary-Vista ($7$ classes).
Top: images from Mapillary validation set;
Middle: ground truth annotations;
Bottom: our model predictions.
   }
\label{fig:exp_qr_2}
\end{figure}

\subsection{Comparison to Prior Art}
\label{subsec:comp_sota}

\subsubsection{EP1}
Table~\ref{table:sc16class} reports semantic segmentation performance of our proposed model trained and evaluated following EP1. 
For a fair comparison with~\cite{tsai2018learning, tsai2019domain, luo2019taking}, we also report results on 13 classes and the standard 16 classes settings.
Our method achieves SOTA performance in EP1 on both $16$ and $13$ classes, outperforming \cite{vu2019dada, lee2018spigan} by large margins.
Now we can identify the major class-specific improvements of our method over the SOTA~\cite{vu2019dada} DADA.
The major gains come from the following classes --
``wall'' ($+12.8\%$), ``motorbike'' ($+10.8\%$), ``bus'' ($+6.9\%$), ``person'' ($+5.5\%$), ``rider'' ($+2.7\%$) and ``car'' ($+2.0\%$).
Moreover, our method shows consistent improvements on classes underrepresented in the target domain:
``motorbike'' ($+10.8\%$), ``pole'' ($+4.0\%$), ``sign'' ($+2.7\%$), and ``bicycle'' ($+1.8\%$).
Fig.~\ref{fig:exp_qr_1} shows the results of the qualitative comparison of our method with DADA~\cite{vu2019dada}.
Note that our model delineates small objects like ``human'', ``bicycle'', and ``motorbike'' more accurately than DADA.

\subsubsection{EP2 and EP3}
Table~\ref{table:scm7class} presents the semantic segmentation results in EP2 and EP3 benchmarks.
The models are evaluated on the Cityscapes and Mapillary validation sets on their common $7$ classes.
We also train and evaluate our model on the $320 \times 640$ resolution to obtain a fair comparison with the reference low-resolution models.
In a similar vein, the proposed method outperforms the prior works in EP2 and EP3 benchmarks for both full- and low-resolution ($640 \times 320$) settings.
We further show in Sec.~\ref{subsec:add_exp_analysis} that our approach achieves state-of-the-art performance without ISL in EP2 and EP3 in both full- and low-resolution settings.
The proposed CTRL coupled with SRH demonstrates consistent improvements over three challenging benchmarks by capitalizing on the inherent semantic and depth correlations.
In EP2 and EP3, our models show noticeable improvements over the state-of-the-art \cite{vu2019dada} with mIoU gains of
$+3.1\%$ (EP2-full-res), $+2.0\%$ (EP2-low-res), $+4.2\%$ (EP3-full-res), $+6.1\%$ (EP3-low-res).
Despite the challenging domain gap between SYNTHIA and Mapillary, our model shows significant improvement ($+6.1\%$) in a low-resolution setting, which suggests
robustness to scale changes.

\begingroup
\setlength{\tabcolsep}{5pt}
\renewcommand{\arraystretch}{1.2}
\begin{table}[t!]
\normalsize
\centering
\caption{
Ablation study of our method from Sec.~\ref{sub:exp_ablation}.
}
\resizebox{\linewidth}{!}{
\begin{tabular}{l cccccccc c}
\toprule
Conf & \rots{SemSup} & \rots{DepSup} & \rots{SRHSup} & \rots{SemAdv} & \rots{DepAvd} & \rots{SRHAdv\quads\quads} & \rots{SL} & \rots{ISL} & mIoU (\%) $\uparrow$   \\
\midrule
$C1$           &  \checkmark       &             &                         &                   &                   &                         &               &           & 30.7 \\
$C2$           &  \checkmark       &  \checkmark &                         &                   &                   &                         &               &           & 35.2 \\
$C3$           &                   &  \checkmark & \checkmark              &                   &                   &                         &               &           & 33.7 \\
$C4$           &  \checkmark       &  \checkmark &  \checkmark             &                   &                   &                         &               &           & 33.1 \\
$C5$           &  \checkmark       &             &                         & \checkmark        &                   &                         &               &           & 40.8 \\
$C6$           &  \checkmark       & \checkmark  &                         & \checkmark        &   \checkmark      &                         &               &           & 40.2 \\
$C7$           &                   & \checkmark  &  \checkmark             &                   &   \checkmark      &     \checkmark          &               &           & 39.5 \\
$C8$           &  \checkmark       & \checkmark  &  \checkmark             &  \checkmark       &   \checkmark      &     \checkmark          &               &           & 42.1 \\
$C9$           &  \checkmark       & \checkmark  &  \checkmark             &  \checkmark       &   \checkmark      &     \checkmark          &  \checkmark   &           & 44.1 \\
$C10$           &  \checkmark       & \checkmark  &                         &  \checkmark       &   \checkmark      &                         &               &\checkmark & 42.8 \\
$C11$          &  \checkmark       & \checkmark  &  \checkmark             &  \checkmark       &   \checkmark      &     \checkmark          &               &\checkmark & \SB{44.9}  \\
\bottomrule
\end{tabular}
}
\label{table:ablation}
\end{table}
\endgroup

\subsection{Ablation Studies} 
\label{sub:exp_ablation}
A comprehensive ablation study is reported in Table \ref{table:ablation}.
We trained $11$ different models, each having a different configuration; these are denoted as $C1, . . ., C11$.
We use the following shortcuts in Table~\ref{table:ablation} to represent different combinations of settings:
``Sem'' -- semantic, ``Dep'' -- depth, ``Sup'' -- supervision, ``Adv'' -- adversarial, and ``Conf'' -- configuration.
Configurations $C1$ to $C4$ denote supervised learning settings without any adversarial training.
These models are trained on the SYNTHIA dataset and evaluated on Cityscapes validation set.
Configurations from $C5$ to $C7$ denote 
different combinations of supervised and adversarial losses on the semantics, depth, and semantics refinement heads.
$C8$ is the proposed model with CTRL, but without ISL.
$C9$ to $C11$ are models trained with SL or ISL with or without SRH. 
$C5$ to $C11$ follow EP1 protocol: SYNTHIA $\rightarrow$ Cityscapes UDA training and evaluation setting.

$C1$ is trained using semantics label supervision without any depth information or adversarial learning. By enabling parts of the model and training procedure, we observed the following tendencies: 
$C2$ \& $C3$ : depth supervision (either direct or through SRH) improves performance;
$C4$: however, adding SRH on top of the depth head in the supervised learning setting does not bring improvements;
$C5$: effectiveness of entropy map domain alignment in semantics feature space \cite{vu2019advent};
$C6$ and $C7$: domain alignment in the depth or refined semantics feature spaces do not bring any further improvements;
$C8$: a combination of depth and SRH with task-specific semantics improves the performance (i.e., our CTRL model);
$C9$: SL brings further improvement but not as good as with our ISL training scheme;
$C10$: emphasizes the improvement over $C6$ with ISL enabled;
$C11$: positive contribution of the SRH towards improving the overall model performance. 
Finally, we achieve state-of-the-art segmentation results (mIoU $44.9\%$) by combining the proposed CTRL, SRH, and ISL (configuration $C11$). 

\begingroup
\setlength{\tabcolsep}{6pt}
\renewcommand{\arraystretch}{1.25}
\begin{table}[t!]
\normalsize
\centering
\caption{
Effectiveness of the joint feature space learned by our method (w/o ISL) for robust domain alignment. Performance in mIoU; legend for ``Ours'' as in Table~\ref{table:scm7class}.
}
\resizebox{\linewidth}{!}{
\begin{tabular}{l @{\quad\quad} c @{\quad\enskip} c @{\quad\enskip} c @{\quad\enskip} c}
\toprule
Model & \makecell{$\text{S}{\rightarrow}\text{C}$ \\ (FR)} & \makecell{$\text{S}{\rightarrow}\text{M}$ \\ (FR)} & \makecell{$\text{S}{\rightarrow}\text{C}$ \\ (LR)} & \makecell{$\text{S}{\rightarrow}\text{M}$ \\ (LR)} \\
\midrule
DADA \cite{vu2019dada}  & 69.2      & 67.6      & 63.4      & 55.8  \\
Ours (best mIoU)        & \SB{70.7} & \SB{67.7} & \SB{64.7} & \SB{58.7}  \\
Ours (confidence)
& \begin{tabular}{@{}r@{}} \SB{70.2} \\[-5pt] {\rpm}\SB{0.6} \end{tabular}  
& \begin{tabular}{@{}r@{}} 66.8 \\[-5pt] {\rpm}0.9 \end{tabular}         
& \begin{tabular}{@{}r@{}} \SB{64.1} \\[-5pt] {\SB{\rpm}}\SB{0.5} \end{tabular}
& \begin{tabular}{@{}r@{}} \SB{58.0} \\[-5pt] {\SB{\rpm}}\SB{0.7} \end{tabular} \\
\bottomrule
\end{tabular}
}
\label{table:comp_dada}
\end{table}
\endgroup

\begingroup
\setlength{\tabcolsep}{6pt}
\renewcommand{\arraystretch}{1.25}
\begin{table}[t!]
\normalsize
\centering
\caption{
Improvement over the state-of-the-art \cite{vu2019dada} in monocular depth estimation. 
The models are trained following SYNTHIA $\rightarrow$ Cityscapes (16 classes) UDA setting w/o ISL and evaluated on the Cityscapes validation set.}
\resizebox{\linewidth}{!}{
\begin{tabular}{lccccccc}
\toprule
Model &  
\rots{$|\text{Rel}|{\downarrow}$} & 
\rots{Rel$^2{\downarrow}$} & 
\rots{RMS{$\downarrow$}} & 
\rots{LRMS{$\downarrow$}} & 
$\delta_1{\uparrow}$ & 
$\delta_2{\uparrow}$ & 
$\delta_3{\uparrow}$ \\
\midrule
DADA~\cite{vu2019dada}  & 0.6      &  10.8       & 17.0       & 4.4    & 0.14      & 0.28      & 0.41  \\
Ours                    & \SB{0.3} &  \SB{6.3}  & \SB{14.8} & \SB{0.6} & \SB{0.30} & \SB{0.58} & \SB{0.77}  \\
\bottomrule
\end{tabular}
}
\label{table:comp_depth}
\end{table}
\endgroup

\subsection{Additional Experimental Analysis}
\label{subsec:add_exp_analysis}
\subsubsection{Effectiveness of the Joint UDA Feature Space}
This section analyzes the effectiveness of joint feature space learned by the CTRL for unsupervised domain alignment. 
We train and evaluate our CTRL model without ISL on two UDA benchmarks:  
(a) EP2: SYNTHIA to Cityscapes $7$ classes (S $\rightarrow$ C) and
(b) EP3: SYNTHIA to Mapillary $7$ classes (S $\rightarrow$ M) in both full- and low-resolution (FR and LR) settings.
In Table \ref{table:comp_dada}, we show the segmentation performance of our model on these four different benchmark settings and compare it against the state-of-the-art DADA model \cite{vu2019dada}.
The proposed CTRL model (w/o ISL) outperforms the DADA model with mIoU gains of
$+1.5\%$, $+0.1\%$, $+1.3\%$, and $+2.9\%$ on all four UDA benchmark settings attesting the effectiveness of the joint feature space learned by the proposed CTRL.

Besides, we train both DADA and our model with ISL and notice improvements in both the models with mIoU $43.5\%$ (DADA) and $44.9\%$ (ours).
The superior quality of the predictions of our model, when used as pseudo labels, provides better supervision to the target semantics; the same can be observed in both our quantitative (Tables~\ref{table:sc16class} and~\ref{table:scm7class}) and qualitative results (Figs.~\ref{fig:exp_qr_1} and~\ref{fig:exp_qr_2}).

\subsubsection{Monocular Depth Estimation Results}
In this section, we show that our model not only improves semantic segmentation but also learns a better representation for monocular depth estimation. This intriguing property is of great importance for multi-task learning. According to \cite{mordan2018revisiting}, paying too much attention to depth
is detrimental to the segmentation performance. Following~\cite{mordan2018revisiting}, DADA~\cite{vu2019dada} uses depth as purely auxiliary supervision. We observed that depth predictions of~\cite{vu2019dada} are noisy (also admitted by the authors), resulting in failure cases. 
We conjecture that a proper architectural design choice coupled with a robust multi-tasking feature representation (encoding task-specific and cross-task relationship) improves both semantics and depth.
In Table~\ref{table:comp_depth}, we report the depth estimation evaluation results on the Cityscapes validation set of our method and compare it against the DADA model~\cite{vu2019dada}. Training and evaluation are done following the EP1 protocol: SYNTHIA $\rightarrow$ Cityscapes (16 classes). We use Cityscapes disparity maps as ground truth depth pseudo-labels for evaluation.
Table~\ref{table:comp_depth} demonstrates a consistent improvement of depth predictions with our method over~\cite{vu2019dada}.
\section{Conclusion}
\label{sec:conclusion}
We proposed a novel approach to semantic segmentation and monocular depth estimation within a UDA context.
The main highlights of this work are:
(1) a Cross-Task Relation Layer (CTRL), which learns a joint feature space for domain alignment; the joint space encodes both task-specific features and cross-task dependencies shown to be useful for UDA;
(2) a semantic refinement head (SRH) aids in learning task correlations;
(3) a depth discretizing technique facilitates learning distinctive relationship between different semantic classes and depth levels;
(4) a simple yet effective iterative self-learning (ISL) scheme further improves the model's performance by capitalizing on the high confident predictions in the target domain.
Our comprehensive experimental analysis demonstrates that the proposed method consistently outperforms prior works on three challenging UDA benchmarks by a large margin.\\ 

\SB{Acknowledgments.} The authors gratefully acknowledge the support by armasuisse. We thank Amazon Activate for EC2 credits and the anonymous reviewers for the valuable feedback and time spent.

\onecolumn
\setlength{\parindent}{0em}

\hsize\textwidth\linewidth
\hsize\toptitlebar 
{
  \centering
  {
    \Large
    \bfseries 
    Learning to Relate Depth and Semantics for Unsupervised Domain Adaptation\\
    Supplementary Materials 
    \par
  }
}
\bottomtitlebar 

\begin{wrapfigure}{R}{0.33\textwidth}
    \vspace{-12pt}
    \includegraphics[width=0.33\textwidth]{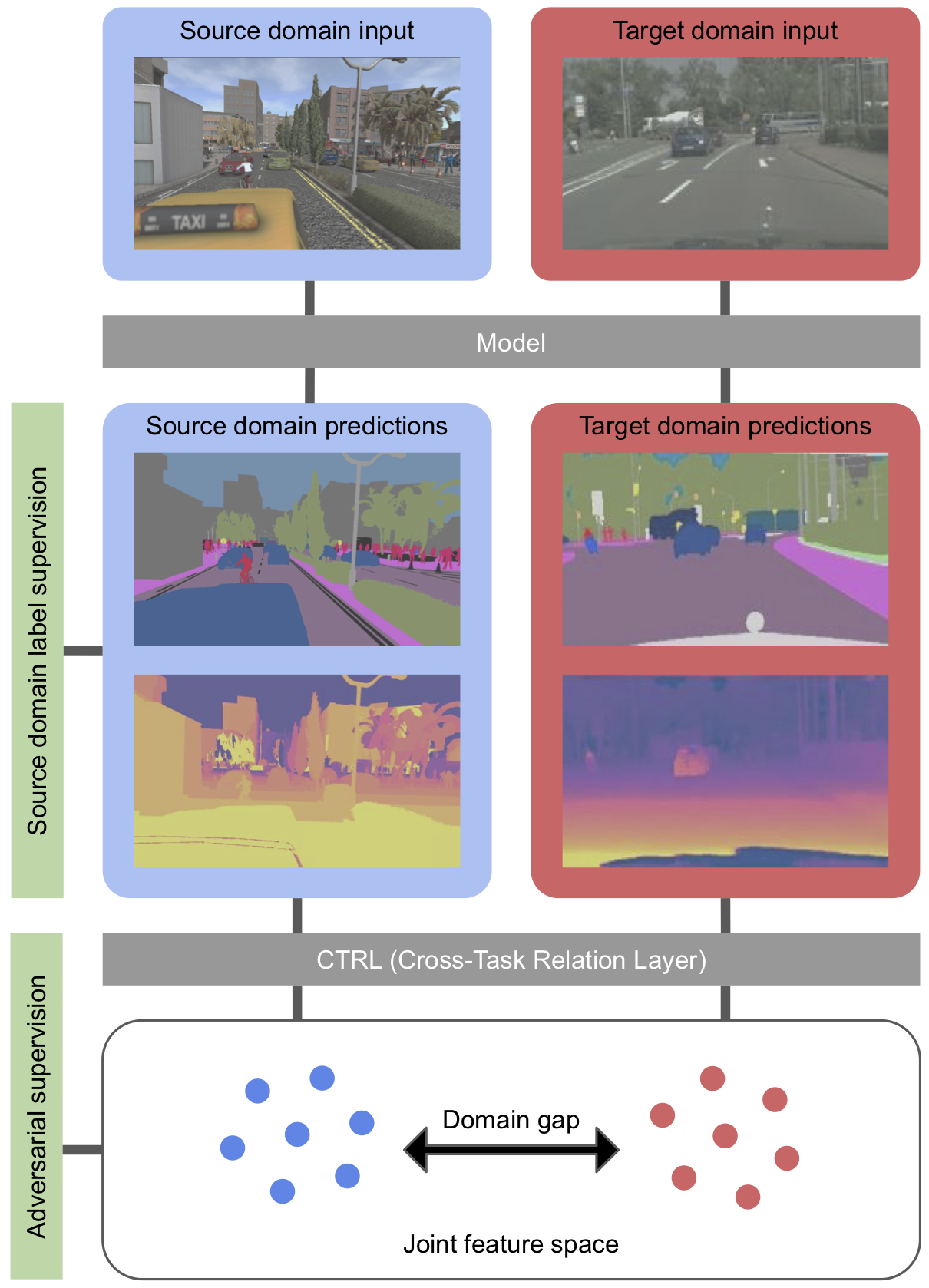}
    \caption{Overview of the UDA setting}
    \label{fig:udarecap}
    \vspace{-30pt}
\end{wrapfigure}
In this document, we provide supplementary materials for our main paper submission. 
First, Sec.~\ref{sec:udarecap} provides a bird-eye view of the assumed UDA setting and how CTRL fits into it.
The main paper reported our experimental results using three standard UDA evaluation protocols (EPs) where the SYNTHIA dataset \cite{ros2016synthia} is used as the synthetic domain.
To demonstrate our proposed method's effectiveness on an entirely new UDA setting, 
in Sec.~\ref{sec:vkitti_to_kitti}, we report semantic segmentation results of our method on a new EP: Virtual KITTI $\rightarrow$ KITTI.
In this setup, we use synthetic Virtual KITTI \cite{gaidon2016virtual} as the source domain and real KITTI \cite{geiger2013vision} as the target domain.
We show that our proposed method consistently outperforms the SOTA DADA method \cite{vu2019dada} 
when evaluated on this new EP with different synthetic and real domains.
In Sec.~\ref{sec:synthia_to_cityscapes}, we present a \mbox{t-SNE}~\cite{tsne} plot comparing our method with \cite{vu2019dada}.
We also share additional qualitative results on SYNTHIA $\rightarrow$ Cityscapes (16 classes).
Sec.~\ref{sec:net_arch_design} details our network design.
To demonstrate that the proposed CTRL is not sensitive to a particular network design (in our case, the residual auxiliary block~\cite{mordan2018revisiting}), 
we train a standard multi-task learning network architecture (i.e., a shared encoder followed by multiple task-specific decoders without any residual auxiliary block) with CTRL and notice a similar improvement trend over the baselines.
The set of experiments and the results are discussed in Sec.~\ref{sec:diff_net_arch}.

\section{Overview of the UDA setting} \label{sec:udarecap}
Unsupervised Domain Adaptation (UDA) aims at training high-performance models with no label supervision on the target domain. As seen in Fig.~\ref{fig:udarecap}, label supervision is applied only on the source domain predictions, whereas tuning the model to perform well on the target domain is the task of adversarial supervision. Since both types of supervision are applied within the same training protocol, adversarial supervision is responsible for teaching the model the specificity of the target domain by means of bridging the domain gap. When dealing with multi-modal predictions, it is crucial to choose the joint feature space subject to adversarial supervision correctly. CTRL provides such rich feature space, which allows training much better models using the same training protocols. This allows us to leverage the abundance of samples in the synthetic source domain and produce high-quality predictions in the real target domain.

\section{Virtual KITTI $\rightarrow$ KITTI} \label{sec:vkitti_to_kitti}
Following \cite{chen2019learning},
we train and evaluate our model on $10$ common classes of Virtual KITTI and KITTI.
In KITTI, the  ground-truth label is only available for the training set; 
thus, we use the official unlabelled test images for domain alignment.
We report the results on the official training set following \cite{chen2019learning}.
The model is trained on the annotated training samples of VKITTI and unannotated samples of KITTI.
For this experiment, we train our model without (w/o) ISL.
Table \ref{table:vk2k10cls} reports the semantic segmentation performance (mIoU\%) of our approach.
Our model outperforms DADA~\cite{vu2019dada}, 
with significant gains coming from the following classes: 
``sign'' ($+8.1\%$), ``pole'' ($+5.7\%$), ``building'' ($+2.7\%$), and ``light'' ($+1.9\%$).
Notably, these classes are practically highly relevant to an autonomous driving scenario.
In Figure \ref{fig:vkitti_to_kitti_qty_results}, 
we present some qualitative results of DADA and our models trained following the new Virtual KITTI $\rightarrow$ KITTI UDA protocol.

\begingroup
\setlength{\tabcolsep}{7pt}
\renewcommand{\arraystretch}{1.25}
\begin{table*}[ht]
\normalsize
\centering
\caption{
Semantic segmentation performance (IoU and mIoU, higher is better, \%) comparison to the prior art.
All models are trained and evaluated using the UDA evaluation protocol Virtual KITTI $\rightarrow$ KITTI.
}
\begin{tabular}{l @{\quad} c @{\quad\quad} cccccccccc @{\quad\quad} c}
\toprule 
&&\multicolumn{11}{c}{VKITTI $\rightarrow$ KITTI (10 classes)} \\
\cmidrule{3-13}
Models & Depth & \rots{road} & \rots{building} & \rots{pole} & \rots{light} & \rots{sign} & \rots{veg} & \rots{terrain} & \rots{sky} & \rots{car} & \rots{truck} & mIoU $\uparrow$ \\
\midrule
Chen \etal \cite{chen2019learning}         & \checkmark & 81.4 & 71.2 & 11.3 & 26.6 & 23.6 & \SB{82.8} & \SB{56.5} & \SB{88.4} & 80.1 & 12.7 & 53.5\\
DADA \cite{vu2019dada}  & \checkmark & \SB{90.9}     & 76.2  & 12.4      & 30.3      & 30.8      & 73.5      & 24.1 & \SB{88.4} & \SB{86.8} & \SB{17.2} & 53.0\\
\midrule
Ours (w/o ISL)          & \checkmark & \SB{90.9} & \SB{78.9} & \SB{18.1} & \SB{32.2} & \SB{38.9} & 73.7      & 22.0      & 88.2      & 86.2      & 16.7 & \SB{54.6} \\
\bottomrule
\end{tabular}
\label{table:vk2k10cls}
\end{table*}
\endgroup

\begin{figure*}[ht!]
\begin{minipage}{\textwidth}
  \begin{minipage}[b]{0.49\textwidth}
    \captionof{table}{Semantic segmentation performance (mIoU) of two variants of the proposed model. Both models outperform DADA~\cite{vu2019dada} attesting the robustness of features learned by the proposed CTRL.}
    \setlength{\tabcolsep}{6pt}
    \normalsize
    \centering
    \resizebox{\linewidth}{!}{
    \begin{tabular}{l @{\quad\quad} c @{\quad} c @{\quad} c}
    \toprule 
    UDA Protocol & DADA & Ours$*$ & Ours \\
    \midrule
    S $\rightarrow$ C 16 cls & 42.6 & 43.7{\rpm}0.2  & \SB{44.3{\rpm}0.6} \\ 
    S $\rightarrow$ C (LR) 7 cls & 63.4 & 63.8{\rpm}0.5  & \SB{64.9{\rpm}0.3} \\ 
    S $\rightarrow$ M (LR) 7 cls & 55.8 & \SB{61.5{\rpm}0.6} & 61.3{\rpm}0.5 \\ 
    S $\rightarrow$ C (FR) 7 cls & 69.2 & \SB{71.3{\rpm}0.5}  & 71.3{\rpm}0.9 \\ 
    S $\rightarrow$ M (FR) 7 cls & 67.6 & 70.1{\rpm}0.5 & \SB{70.9{\rpm}0.7} \\ 
    \bottomrule
    \end{tabular}
    }
    \label{table:net_arch_comp}
  \end{minipage}
  \hfill
  \begin{minipage}[b]{0.49\textwidth}
    \centering
    \includegraphics[width=0.88\linewidth]{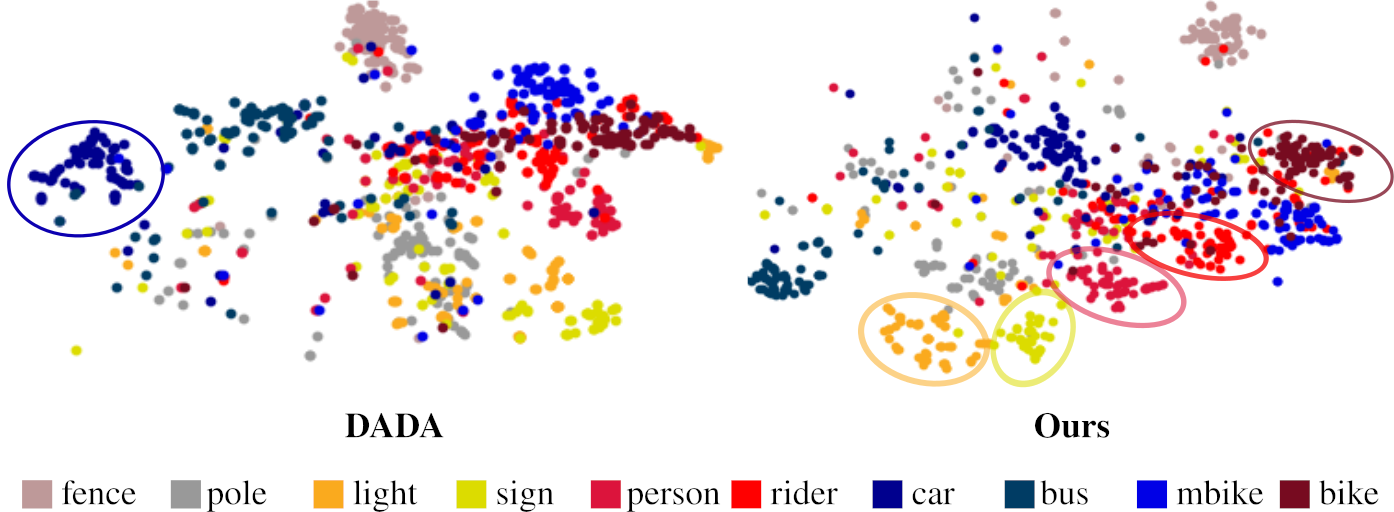}
    \captionof{figure}{t-SNE comparison of features learned by DADA~\cite{vu2019dada} and CTRL. It leads to more structured feature space and better class separation in the target domain. Circled classes have a better separation than the other method.}
    \label{fig:t-SNE}
  \end{minipage}
\end{minipage}
\end{figure*}

\begin{figure*}[ht!]
\begin{center}
   \includegraphics[width=1.0\linewidth]{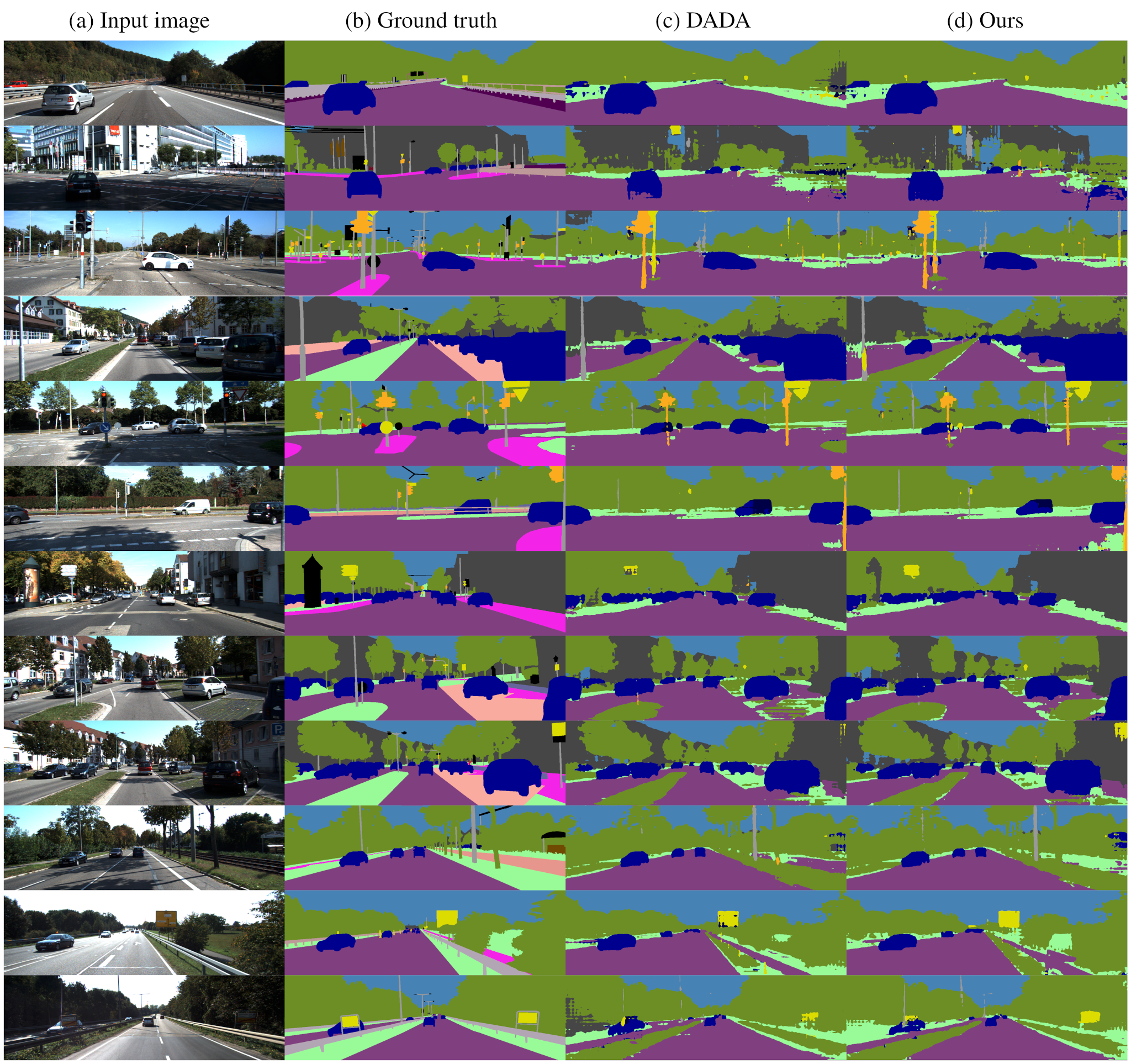}
\end{center}
\caption{
Qualitative semantic segmentation results with VKITTI $\rightarrow$ KITTI (10 classes) UDA evaluation protocol.
(a) Input image from the target domain KITTI;
(b) ground truth annotations;
(c) DADA~\cite{vu2019dada} predictions;
(d) our model predictions.
We follow the color encoding scheme of Cityscapes to colorize the label maps.
}
\label{fig:vkitti_to_kitti_qty_results}
\end{figure*}

\section{SYNTHIA $\rightarrow$ Cityscapes} \label{sec:synthia_to_cityscapes}
\begin{figure*}[ht!]
\begin{center}
   \includegraphics[width=1.0\linewidth]{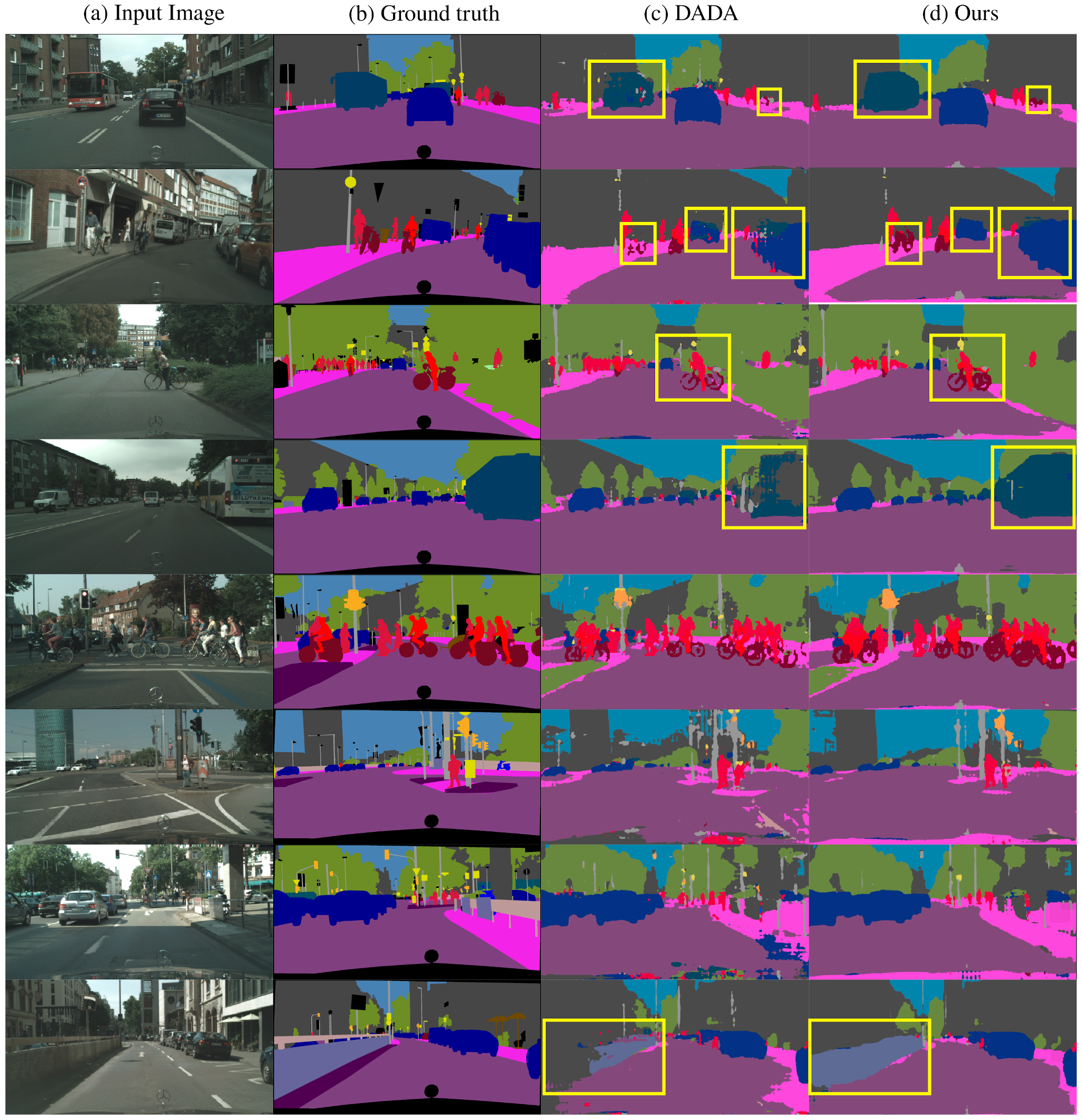}
\end{center}
\caption{
Qualitative semantic segmentation results with EP1: SYNTHIA $\rightarrow$ Cityscapes (16 classes).
(a) Images from Cityscapes validation set;
(b) ground truth annotations;
(c) DADA~\cite{vu2019dada} predictions;
(d) our model predictions.
Our method demonstrates notable improvements over \cite{vu2019dada} on ``bus'', ``person'', and ``bicycle'' classes as highlighted using the yellow boxes.
}
\label{fig:synthia_to_cityscapes_16cls_img1}
\end{figure*}

\begin{figure*}[ht!]
\begin{center}
   \includegraphics[width=1.0\linewidth]{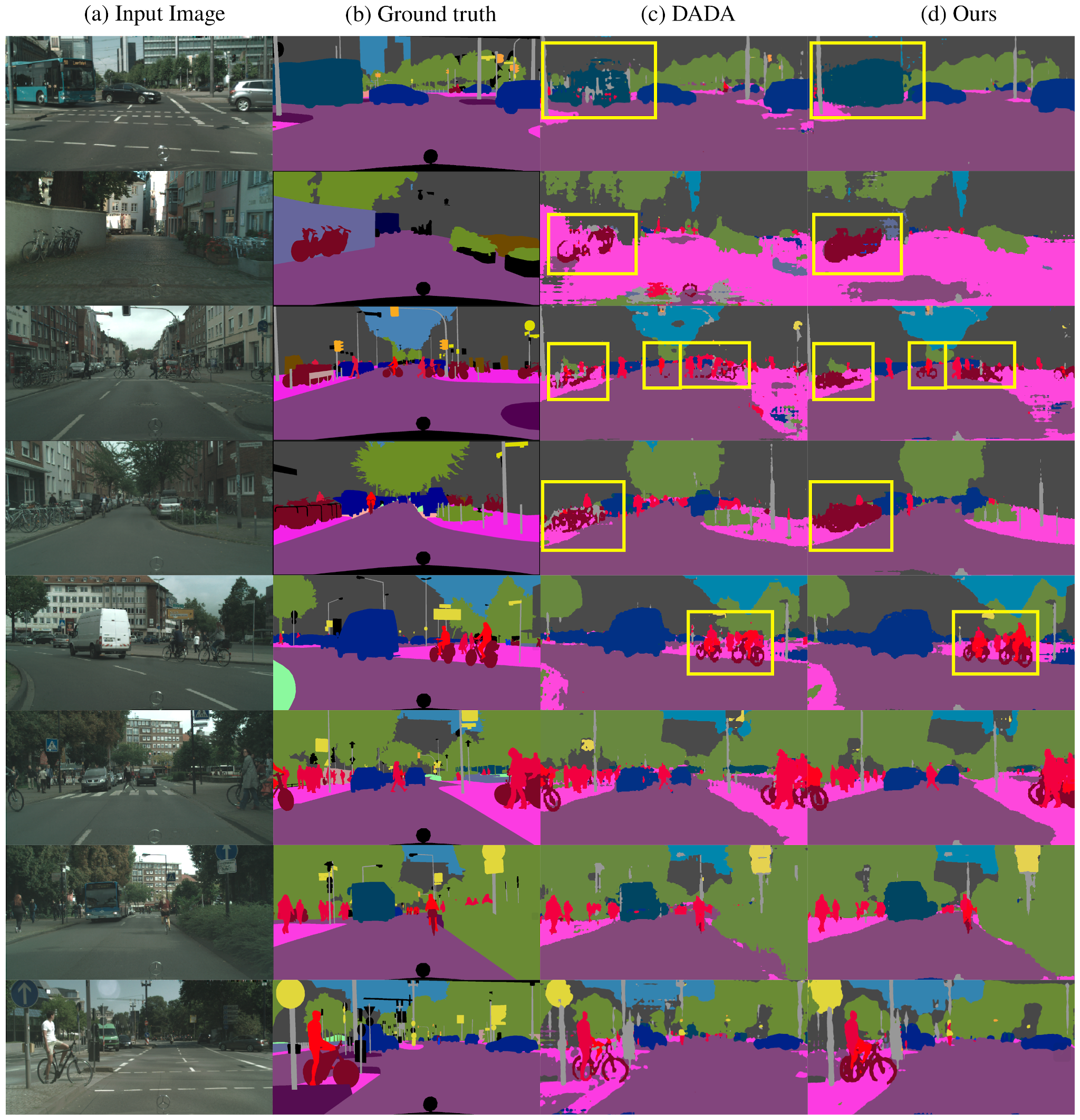}
\end{center}
\caption{
Qualitative semantic segmentation results with EP1: SYNTHIA $\rightarrow$ Cityscapes (16 classes).
(a) Images from Cityscapes validation set;
(b) ground truth annotations;
(c) DADA~\cite{vu2019dada} predictions;
(d) our model predictions.
Our method demonstrates notable improvements over \cite{vu2019dada} on ``bus'', ``person'', and ``bicycle'' classes as highlighted using the yellow boxes.
}
\label{fig:synthia_to_cityscapes_16cls_img2}
\end{figure*}

\begin{figure*}[ht!]
\begin{center}
   \includegraphics[width=1.0\linewidth]{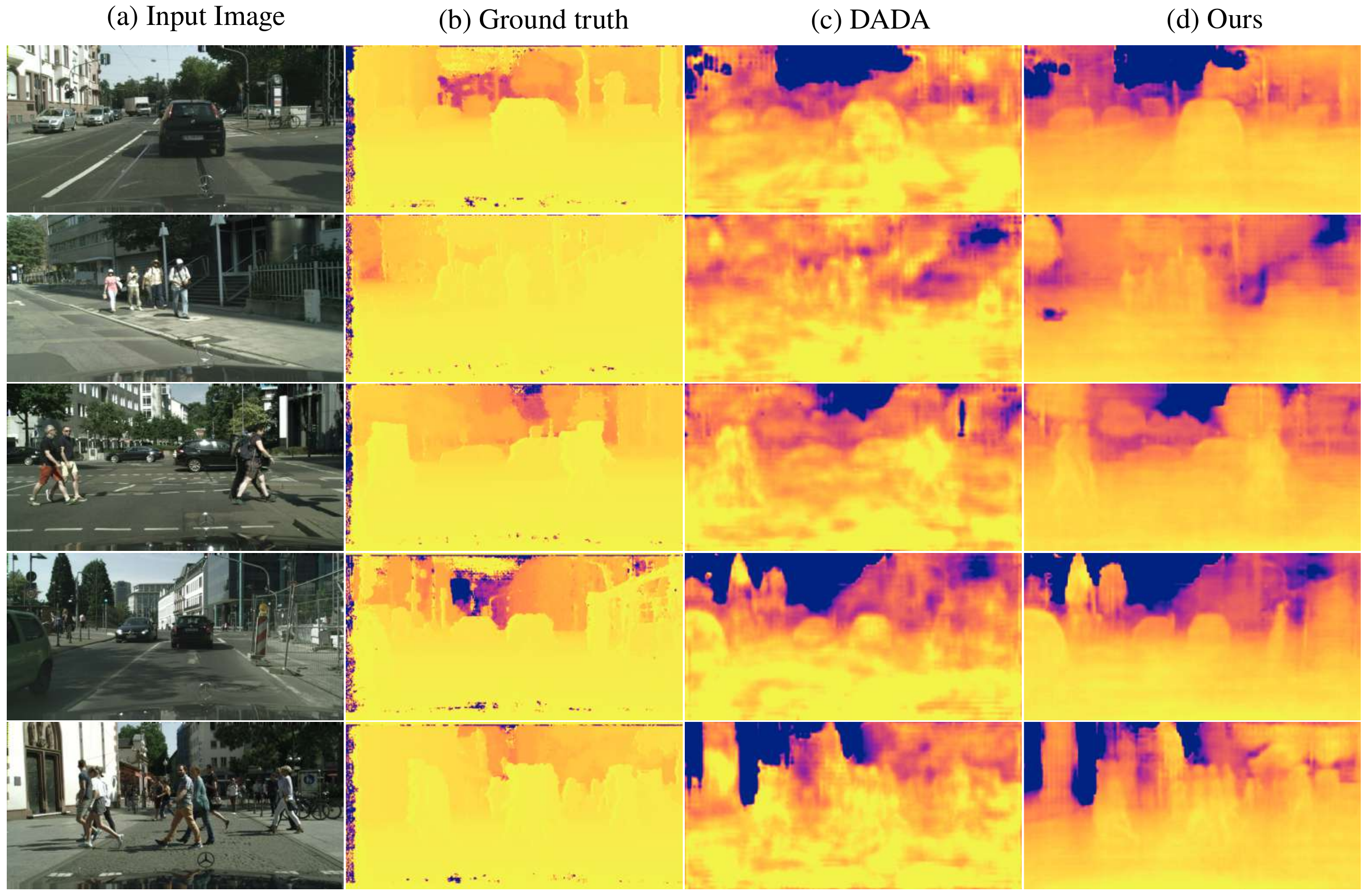}
\end{center}
\caption{
Qualitative monocular depth estimation results with EP1: SYNTHIA $\rightarrow$ Cityscapes (16 classes).
(a) Images from Cityscapes validation set;
(b) ground truth annotations;
(c) DADA~\cite{vu2019dada} predictions;
(d) our model predictions.
}
\label{fig:synthia_to_cityscapes_16cls_img3}
\end{figure*}

This section presents a t-SNE~\cite{tsne} plot of the feature embeddings learned by the proposed model guided by CTRL, and \cite{vu2019dada}.
Fig.~\ref{fig:t-SNE} shows 10 top-scoring classes of each method; distinct classes are circled. As can be seen from the figure, CTRL leads to more structured feature space, which concurs with our analysis of the main paper.
Both models are trained and evaluated following the UDA protocol SYNTHIA $\rightarrow$ Cityscapes (16 classes).
Furthermore, we present additional qualitative results of our model for semantic segmentation and monocular depth estimation.
Figures \ref{fig:synthia_to_cityscapes_16cls_img1},~\ref{fig:synthia_to_cityscapes_16cls_img2} 
show the results of the qualitative comparison of our method with \cite{vu2019dada}. 
Note that our proposed method has higher spatial acuity in delineating small objects like “human”, “bicycle”, and “person” compared to \cite{vu2019dada}.
Figure \ref{fig:synthia_to_cityscapes_16cls_img3} shows some qualitative monocular depth estimation results.

\section{Network Architecture Design} \label{sec:net_arch_design}
The shared part of the semantic and depth prediction network $\mathcal{F}_e$ consists of a ResNet-101 backbone and a decoder.
The decoder consists of four convolutional layers, each followed by a Rectified Linear Unit (ReLU). 
The decoder outputs a feature map that is shared among both semantics and depth heads.
This shared feature map is fed forward to the respective semantic segmentation, monocular depth estimation, and semantics refinement heads.
For the task-specific and task-refinement heads, we use Atrous Spatial Pyramid Pooling (ASPP) with sampling rates $[6, 12, 18, 24]$ and the Deeplab-V2~\cite{chen2017deeplab} architecture.
Our DC-GAN~\cite{radford2015unsupervised} based domain discriminator takes as input a feature map with channel dimension $2 \times C + K$, where $C$ is the number of semantic classes, and $K$ is the number of depth levels.

\section{Robustness to Different Network Design} \label{sec:diff_net_arch}
Our proposed model adopts the residual auxiliary block \cite{mordan2018revisiting} (as in \cite{vu2019dada}),
which was originally proposed to tackle a particular MTL setup where the objective was to improve one primary task by leveraging several other auxiliary tasks.
However, unlike \cite{vu2019dada} which doesn't have any decoder for depth, we introduce a DeepLabV2 decoder for depth estimation to improve both task performances. Our qualitative and quantitative experimental results show an improvement of depth estimation performance over \cite{vu2019dada}. 
Furthermore, we are interested to see the proposed model's performance when used with a standard MTL architecture (a common encoder followed by multiple task-specific decoders without any residual auxiliary blocks). 
To this end, we make necessary changes to our existing network design to have a standard MTL network design.
We then train it following UDA protocols. The details of our experimental analysis are given below.

For the standard MTL model (denoted as ``Ours*'' in Table  \ref{table:net_arch_comp}), the depth head is placed after the shared feature extractor $\mathcal{F}_e$. 
The shared feature extractor consists of a ResNet backbone and decoder network (see Fig. 2).
For the second model with residual auxiliary block (denoted as ``Ours''), we positioned the depth head after the decoder's third convolutional layer.
The semantic segmentation performance of these two variants of the proposed model is shown in Table \ref{table:net_arch_comp}. 
Both models are evaluated on the five different UDA protocols and outperform state-of-the-art DADA~\cite{vu2019dada} results. 
The results show that our proposed CTRL is not sensitive to architectural changes and can be used with standard encoder-decoder MTL frameworks.
Our findings may be found beneficial for the domain-adaptive MTL community, e.g., in answering a question whether \it{learning additional complementary tasks (surface normals, instance segmentation) performs domain alignment}.

{
  \small
  \balance
  \bibliographystyle{ieee_fullname}
  \bibliography{egbib}
}

\end{document}